# Large-Scale Data Mining of Rapid Residue Detection Assay Data From HTML and PDF Documents: Improving Data Access and Visualization for Veterinarians

Majid Jaberi-Douraki[1,2,3*], Soudabeh Taghian Dinani[1,3,4], Nuwan Indika Millagaha Gedara[1,3,5], Xuan Xu[1,2,3], Emily Richards[6], Fiona Maunsell[7], Nader Zad[1,2,8] and Lisa A. Tell[6]

[1] DATA Consortium, www.1DATA.life, Olathe, KS, United States, [2] 1DATA, Kansas State University Olathe, Olathe, KS, United States, [3] Department of Mathematics, Kansas State University, Manhattan, KS, United States, [4] Department of Computer Science, Kansas State University, Manhattan, KS, United States, [5] Department of Business Economics, University of Colombo, Colombo, Sri Lanka, [6] Department of Medicine and Epidemiology, School of Veterinary Medicine, University of California, Davis, Davis, CA, United States, [7] Large Animal Clinical Sciences, College of Veterinary Medicine, University of Florida, Gainesville, FL, United States, [8] Department of Civil Engineering, Kansas State University, Manhattan, KS, United States

Extra-label drug use in food animal medicine is authorized by the US Animal Medicinal Drug Use Clarification Act (AMDUCA), and estimated withdrawal intervals are based on published scientific pharmacokinetic data. Occasionally there is a paucity of scientific data on which to base a withdrawal interval or a large number of animals being treated, driving the need to test for drug residues. Rapid assay commercial farm-side tests are essential for monitoring drug residues in animal products to protect human health. Active ingredients, sensitivity, matrices, and species that have been evaluated for commercial rapid assay tests are typically reported on manufacturers' websites or in PDF documents that are available to consumers but may require a special access request. Additionally, this information is not always correlated with FDA-approved tolerances. Furthermore, parameter changes for these tests can be very challenging to regularly identify, especially those listed on websites or in documents that are not publicly available. Therefore, artificial intelligence plays a critical role in efficiently extracting the data and ensure current information. Extracting tables from PDF and HTML documents has been investigated both by academia and commercial tool builders. Research in text mining of such documents has become a widespread yet challenging arena in implementing natural language programming. However, techniques of extracting tables are still in their infancy and being investigated and improved by researchers. In this study, we developed and evaluated a data-mining method for automatically extracting rapid assay data from electronic documents. Our automatic electronic data extraction method includes a software package module, a developed pattern recognition tool, and a data mining engine. Assay details were provided by several commercial entities that produce these rapid drug residue assay tests. During this study, we developed a real-time conversion system and method for reflowing contents in these files for accessibility practice and research data mining. Embedded information was extracted using an AI technology for text extraction and text mining to convert to structured formats. These data were then made available to veterinarians and producers via an online interface, allowing interactive searching and also presenting the commercial test assay parameters in reference to FDA-approved tolerances.

## INTRODUCTION

Drug residue testing is an essential tool to ensure that animal products intended for human consumption are free of violative residues (1). These tests are most commonly used to test the milk before it is added to the bulk tank but some tests can also be used to evaluate other matrices (urine, serum, eggs, and honey) to ensure that drug concentrations are below the tolerance after the withdrawal interval has been observed. Tolerance refers to the maximum acceptable level of a chemical residue present in food products from an exposed/treated animal, which is determined by the Food and Drug Administration (2). Rapid quantitative drug detection has largely been applied to help minimize drug residue risks and maintain milk quality (3–5). To use these tests, information for different commodities including cow, swine, goat, sheep, camel, horse, and buffalo and matrices including serum, urine, milk, and honey is necessary and have been documented for approximately 100 different rapid test assays (1, 3, 6). Contents of these documents are mostly published in semi or unstructured portable document format (PDF) files or hypertext markup language (HTML) documents which do not allow for interactive searching and easy comparison to tolerance limits.

Extracting tables from PDF files and HTML documents has been investigated both by academia and commercial tool builders (7). However, the techniques of extracting tables from PDF and HTML are still open and new techniques including dynamic and automatic statistical text mining are currently being studied by the researchers (8–10). Besides, techniques for extracting tables with HTML in Web pages (11–13) are different than identifying tables from scanned documents (14, 15). Moreover, PDF documents have no specific table markups; text-based PDF documents need parsing procedures that are modified for processing the raw PDF format (16, 17). On the other hand, table extraction techniques tend to be devised concerning the application context. For instance, the Web Table Corpora focuses on knowledge base construction (18) and deals with matching table instances to concepts in DBpedia or Wikidata (19). In particular, the text-based PDF processing work is impacted by a lack of standardized schema for PDF parsers, leading to limited applicability.

The HTML markups for tables are used for page layout, and only a subset of the "tables" actually contain tabular data (20). However, it also focuses on separating table data from layout instructions and provides large-scale Web table data classified from general-purpose Web crawl data (18). The processing of Web table can be categorized into two categories: table search and knowledge base construction, one of the most commonly used approaches which is a keyword-based technique that ranks extracted tables based on table content (21). Next, matching tables can be applied to complete or extend a table itself (22). Thus, these techniques use a machine learning approach to leverage the relationships amongst rows and columns.

In the current study, we developed a data-mining method to automatically extract commercial rapid assay data from electronic documents to ensure accurate data and so the data could be searched using an interactive interface. Our automatic electronic data extraction system comprises a software package module, a developed pattern recognition tool, and a data mining engine. Data was harvested from online websites or generously supplied by manufacturers for the available commercial assays. Some of the data are sporadically reported online but most were published in semi or unstructured PDF files or HTML documents. We developed a real-time conversion system and method for reflowing contents in these PDF and HTML files for accessibility practice and quantitative research data mining. Embedded information was extracted using an artificial intelligence (AI) technology for text extraction and text mining to convert to structured formats. The data were usually hidden in the main text and mostly in the form of a tabulated summary.

## MATERIALS AND METHODS

There are two main methodological steps in this study, one is to extract the desired information from specified documents, and the second is to update the previously available information based on these new values. In the following, we show a summary of our workflow integrated by machine learning where first some preprocessing points are presented, and then these steps are provided in detail for data extraction (**Figure 1**).

### Real-Time Data Collection via PDF and Webpage Parsing

Two formats of documents extracted from Web sources were virtually imported for this study: files (34 tables out of 60 relevant pdf files, there were also over 180 more PDFs either did not have tables or extracted tables did not have relevant fields) were produced and made available online by seven manufacturers in PDF formats while other tables were presented by HTML documents on Web sources. Automating the extraction of data from structured HTML tables was implemented with easy-to-use and effective web data extraction techniques. However, the contents of these reports are mostly published in unstructured PDF files which were also the main challenge in our research settings for extracting information.

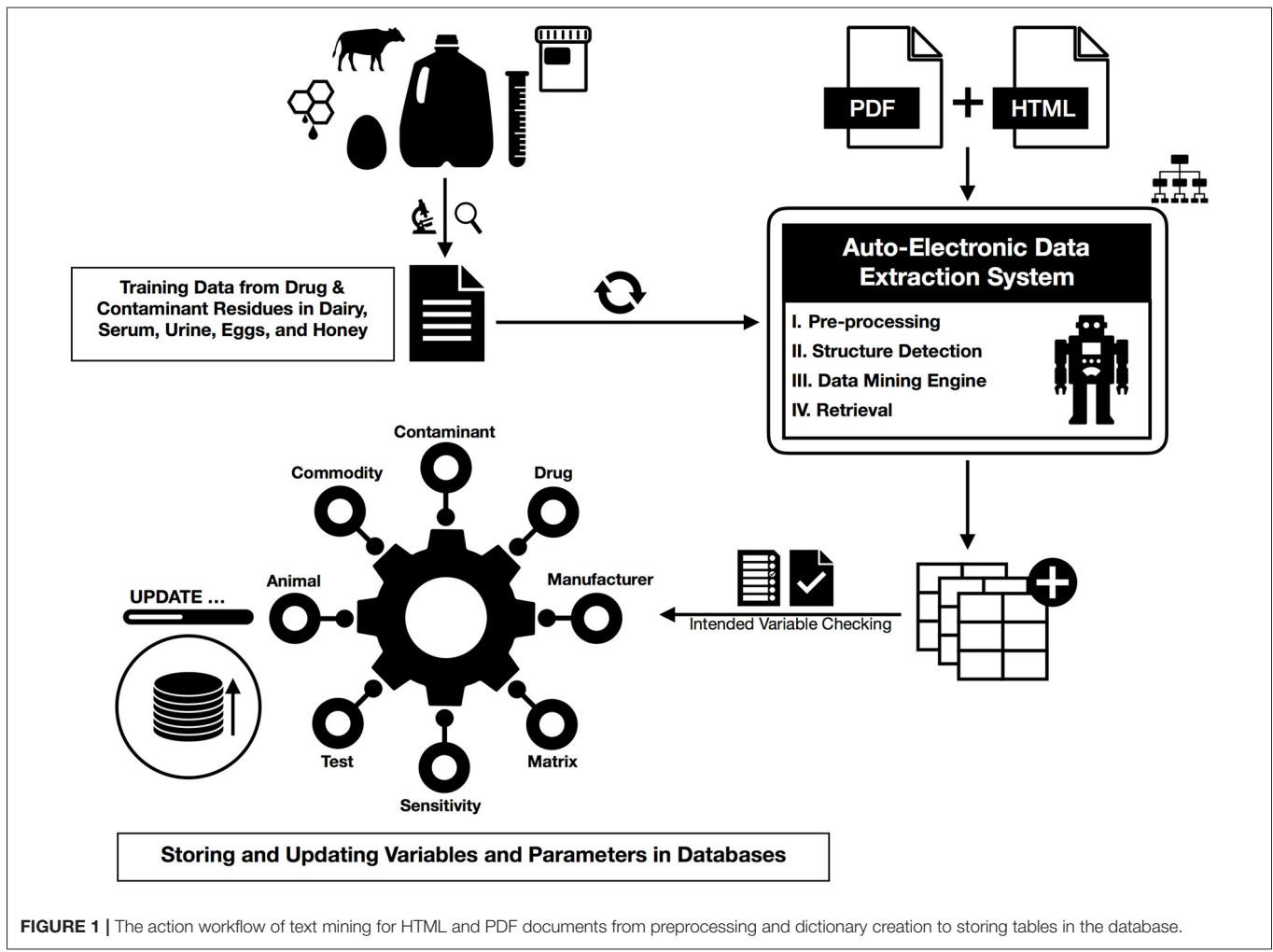

FIGURE 1 | The action workflow of text mining for HTML and PDF documents from preprocessing and dictionary creation to storing tables in the database.

In our work, the text-mining and information-retrieval models for rapid test assays were trained to curate data from manufacturer and producer manuals for drug and contaminant residues that may be found in dairy products, serum, urine, eggs, and honey (1, 3, 6). These tests aim to detect drug or contaminant residues at or below established tolerances or maximum residue limits. A maximum residue limit (MRL) is similar to tolerance in that it is the maximum limit of a chemical residue acceptable in food products obtained from an exposed or treated animal, however, it is determined by the European Medicines Agency.

In our real-time text mining, the data collected from residue tests in veterinary medicine was 2-fold. In the first step, values or desired fields in a search query corresponding to the intended variables such as sensitivity, commodity, matrix, drug or contaminant, animal, manufacturer, or test are automatically extracted from the tables obtained from Web sources including PDFs and HTML documents. In the second step, these extracted data elements are compared to the previously obtained fields in our datasheets. If no fields match the new query, it means that we have received new data updated by a manufacturer that had not been previously reported and a new row will be added to the previous table (along with the other fields). On the other hand, if the fields other than sensitivity are the same (drug/contaminant name, test name, matrix, and test type) and a change was received in the sensitivity value, the new value will be updated and stored in the datasheet. Following data mining, tolerance information collected via the electronic Code of Federal Regulations (2) was manually added to each corresponding drug or contaminant line in the datasheet to allow for a visual comparison of test sensitivity to applicable tolerance, which may help determine test suitability. The datasheet was then uploaded into an interactive, searchable interface online that allows veterinarians or producers to query.

## Information Extraction Tools and Software for PDF and HTML Documents

For PDF files, *PdfFileReader* from *PyPDF2*, a Python exclusive PDF module was used to scan PDF files (Python Software

Foundation. Python Language Reference, version 3.7. Available at http://www.python.org). Then *read_pdf* from the *tabula* module was imposed to detect tables on an individual page from a single PDF file. Drug names, sources, URL, and contact information were extracted based on the preferred drug list created beforehand and saved into separate excel files.

For HTML websites, the *webdriver* function from the Python module *selenium* was used to navigate to a webpage source accompanied by a specific URL address. The HTML and XML document parser, *BeatifulSoup*, from Python module BS4 was implemented to parse the source page; multimodal texts of Web page sources were extracted as *BeautifulSoup* object. *Pandas read_html* as a data analysis miner and manipulation and powerful web scraping tool for URL protocols was used to harvest data from HTML tables.

## Regular Expression Learning for Information Extraction

To briefly explain the high-throughput regular-expression pattern matching method, we have implemented some similarity methods from Regular Expression techniques in our Python codes that help identify patterns to match similar/missing characters or fields and create a synonym/dictionary table (23, 24). The textual semantic similarity measures based on web intelligence of an ensemble of keywords were processed by using regular expressions when cleaning and merging information of sensitivity, drugs, or tests from different sources.

## The Information Available in Each Field

Below is a table of all necessary fields used for parsing webpage sources when a query is submitted to extract data from PDF or HTML documents.

# RESULTS
## Preprocessing

Initially, we started the text mining model based on an original Microsoft Excel file including information previously obtained from text mining and manual curation of rapid assay tests of PDF files and HTML websites (1, 3, 6). We faced multiple technical issues: (1) the drugs' names and bioassay test names in this file did not follow any specific standard, thus were not consistent for data extraction. For example, assay methods (e.g., sequential, competitive, or quantitative) were often written in the same field as the drug name, comma-separated in front of the field, or ahead of a test name. (2) Formulations of the same active ingredient with different generic or trade names created confusion in collecting data (for instance, benzylpenicillin procaine and benzathine penicillin). Similarly, some rapid diagnostic assay test names were presented differently in each document that might have referred to the same test (for instance, Charm KIS, Charm Kidney Inhibition Swab, and KIS referring to the same test).

To deal with the first problem for the data curation, we implemented a simple unsupervised learning algorithm to identify synonyms, based on the data amassed by implementing a web search engine for specific tables from multiple documents. The algorithm used pointwise mutual information (PMI) and information retrieval (IR) to measure the similarity of pairs of words by reporting the experiment types, if exist in any instance, to a separate column (25, 26). For the second problem, the same method was used to discover synonymy in extracting semantic information which has been of high importance in information retrieval and automatic indexing. For this purpose, the most frequent names for each drug, matrix, and test were used for synonyms based on frequency in the file. Also, readings in information retrieval were stored in a Microsoft Excel

| Field | Available information |
|---|---|
| Test | 2,4-D RaPID Assay, Atrazine RaPID Assay, Benomyl RaPID Assay, BetaStar Advanced for Beta-Lactams, BetaStar Advanced for Tetracycline, Beta Star 4D, BetaStar 4D Beta-Lactam, Tetracycline, Streptomycin, Chloramphenicol, BetaStar S for Sulfonamides, BetaStar for Quinolone, Charm II Aflatoxin, Charm II Amphenicol, Charm II Beta-Lactam, Charm II Chloramphenicol, Charm II Cloxacillin, Charm II Gentamicin and Neomycin, Charm II Gentamicin and Streptomycin, Charm II Macrolide, Charm II Novobiocin, Charm II Streptomycin, Charm II Sulfonamide, Charm II Tetracycline, Charm 3 SL3 Beta-Lactam, Charm B. stearothermophilus Tablet Disc Assay, Charm Beta-Lactam 30 s, Charm Blue Yellow II, Charm Cowside II, Charm Enroflox, Charm Flunixin and Beta-Lactam, Charm Gentamicin, Charm HPLC Receptogram, Charm KIS, Charm MRL Beta-Lactam, Charm MRL Beta-Lactam 1-min, Charm MRL Beta-Lactam 3-min, Charm MRL Beta-Lactam and Tetracycline 2-min, Charm MRL Beta-Lactam and Tetracycline, Charm MRL Beta-Lactam and RF Tetracycline |
| | 2-min, Charm Quad, Charm Quad 1, Charm Quad 2, Charm Quad 3, Charm Quinolone, Charm ROSA Amphenicol, Charm ROSA Chloramphenicol, Charm ROSA MRL Aflatoxin, Charm ROSA SL Aflatoxin, Charm ROSA Macrolide, Charm ROSA Neomycin and Streptomycin, Charm ROSA Pirlimycin, Charm ROSA Streptomycin, Charm ROSA Sulfa, Charm ROSA Tetracycline, Charm SL Beta-Lactam, Charm Streptomycin, Charm Tetracycline, Charm TRIO, Delvotest BLF, Delvotest P 5 Pack, Delvotest P/ Delvotest P Mini, Delvotest SP-NT, Delvotest T, Eclipse 3G, Meatsafe B-Lactam, Meatsafe Gentamicin Strip, New Snap Beta-Lactam, Penzyme Milk Test, PremiTest, Reveal for Aflatoxin M1, SNAP Aflatoxin M1, SNAP AFM1, SNAP Beta-Lactam ST, SNAP Beta-Lactam ST Plus, SNAP Duo ST Plus, SNAP Gentamicin, SNAP NBL, SNAP Sulfamethazine, SNAP Tetracycline, SNAP TRIO Japan, SNAPduo ST Plus, Veratox for Enrofloxacin, Veratox for Florfenicol, Veratox for Fluoroquinolone, Veratox for Gentamicin, Veratox for Neomycin, Veratox for Oxytetracycline, Veratox for Sulfonamides, Veratox for Tetracycline, Veratox for Tylosin. |
| Drug/Contaminant | 2,4-D, Aflatoxin M1, Amoxicillin, Ampicillin, Atrazine, Bacitracin, Cefoperazone, Cefquinome, Ceftiofur, Cephalexin, Cephapirin, Clindamycin, Cloxacillin, Chloramphenicol D, Chlortetracycline, Danofloxacin, Dapsone, Dicloxacillin, Dihydrostreptomycin, Enrofloxacin, Erythromycin, Florfenicol, Flunixin, Gentamicin, Hetacillin, Kanamycin, Lincomycin, Neomycin, Novobiocin, Oxacillin, Oxytetracycline, Penicillin, Pirlimycin, Polymixin B, Rifaximin, Spectinomycin, Streptomycin, Sulfachlorpyridazine, Sulfadiazine, Sulfadimethoxine, Sulfadoxine, Sulfaethoxypyridazine, Sulfamerazine, Sulfamethazine, Sulfamethizole, Sulfamethoxazole, Sulfanilamide, Sulfapyridine, Sulfaquinoxaline, Sulfathiazole, Tetracycline, Thiamphenicol, Tilmicosin, Trimethoprim, Tulathromycin, Tylosin. |
| Animal | Buffalo, Camel, Cattle, Chicken, Honeybees, Horse, Sheep, Swine, Turkey, Various fish/shellfish species |
| Manufacturer | Charm Sciences, Inc; DSM Food Specialties USA, Inc; IDEXX Laboratories, Inc; Neogen Corporation; Silver Lake Research Corporation; Strategic Diagnostics, Inc; ZEU-Immunotec |
| Matrix | Serum, Urine, Milk, Honey, Egg |
| Type | Sequential, Competitive, Quantitative |

TABLE 1 | Intended keywords to be checked in the extracted tables.

| Parameter type | Ensemble of keywords to be used for data miner |
|---|---|
| Desired matrix | "\Wmilk\W", "\Whoney.?\W", "\Wserum\W\|\WSera\W", "\Wtissue.?\W", "\Wkidney\W", "\Wdairy\W", "\Waquaculture\W", "\Wurine\W" |
| Desired field | "\Wsafety\W", "\Wconcentration for positive ppb\W", "\Wcharms\|\W", "\Wsensitivity\W", "\Wtest sensitivity\W", "\Wconcentration\W", "\Wpositive concentration\W", "\Wdetection level\W", "\Wdetection range\W" |
| Unit of the field | "\Wppb\W", "\Wppm\W" |

TABLE 2 | Dictionary for synonyms or corresponding names considered for each field.

| Main field name | Other names used in the original documents |
|---|---|
| "Drug" | "\WActive.?ingredient.?\W", "\WResidues.?detected\W", "\WDetected.?residues\W", "\WAntimicrobial.?drug\W", "\WAntimicrobial.?\W", "\WAntimicrobial.?agent.?\W", "\WBeta.?lactam drug.?\W", "\WBeta lactams\W", "\WTetracycline.?\W", "\WQuinolone.?drug.?\W" |
| "Sensitivity" | "\WConcentration for positive ppb\W", "\WCharm.?sl\W", "\WTest.?Sensitivity\W", "\WConcentration\W", "\WPositive Concentration\W", "\WDetection level\W", "\WDetection range\W", "\WDetection range ppb\W" |
| "Test" | "\WTest\W", "\WTest Name\W" |
| "Matrix" | "\WMatrix\W", "\WSpecimen\W" |
| "MRL" | "\WCODEX\W", "\WTechnical Regulation\W", "\WMRPL\W", "\WFederation\W", "\WRegulation\W", "\WImport Regulation\W", "\WMRL.?ppb\W" |
| "Tolerance" | "\WAction Level\W", "\WSafe level\W", "\WSafety level\W" |

*Names are followed by some regular expressions to ensure correct field extractions.*

spreadsheet file for the drug names and other fields including the matrix, manufacturer, or test names. Briefly, the most frequent name was reported in the first column and other names for the corresponding drug, matrix, or test were reported in the other columns in front of that (each name in one column). These two files were further cleaned if changes for drug names or other fields were observed while extracting data tables.

## Desired Information From Structured or Unstructured Documents

Below we review multiple cases to extract data from tables. For these cases, it is required to check if the keywords determined important in the real-time data collection via PDF and webpage parsing are clearly characterized in the extracted tables provided that any data are available. The necessary information is presented by their types and ensemble of keyword extraction from a single document using word co-occurrence statistical information in **Table 1**. Here each item is supported by a regular expression to make sure each keyword is not part of another word to misclassify the keyword. For instance, urine can be found in purines or tissue as part of intertissued. Using those regular expressions, we avoid the misclassification of a keyword preceded or followed by the other prefixes, suffixes, or words. Here using the regular expression matching "\W" ensures no letter or numbers will be part of a keyword and the bar "|" acts as an OR operator for the regular expression search algorithm.

However, each field can be presented differently as authors may use incoherent terminologies based on their backgrounds or other international standards. To make all the field names consistent and create a synonyms table, the field names are compared to the list of names for each field, and they are replaced by a simple name as the main field name (given in **Table 2**). Some other fields such as MRL or Tolerance may also be considered in the data extraction process since these fields are available for retrieval in some documents and tables, and but not all the time. This creates repeated columns and we are required to deal with such cases since the names of repeated columns should be consistent. Here we similarly attempted to authenticate the input string of each field using regular expression matching to cover more cases in our queries.

## Extracting Semistructured Information From the Web

As mentioned before, data mining of information for rapid drug residue assays is an essential tool in veterinary medicine with source information ranging from tables in PDF files to HTML text which is presented on the Web. The data and information-retrieval model for rapid assays were obtained from the dairy products for groups of antibiotics including beta-lactams, tetracyclines, aminoglycosides, and sulfonamides from the website www.idexx.com/en/milk/ (6). Therefore, using our trained model based on the Python packages of *requests* and *BeautifulSoup*, all the rapid assay URL links for dairy tests are parsed and automatically examined for potentially available tables on each page. Below we presented an example of adaptable parsing of real-time data extracting. When the query pinpoints the above-said keywords in the extracted table as presented in **Table 1**, the desired fields along with the source link of the table are collected and stored in an Excel file. For this particular example, the matrix is found as "milk" from this link and then it is added as a new field to the stored file. In addition, we can also identify the title of each dairy test page by checking the HTML file and then use it as the "test name" corresponding to each table; this requires us to create another new field as well.

In **Figure 2**, we found the dairy test for the "SNAP NBL Test" which "detects beta-lactam residues at or below the U.S. FDA-established tolerance/safe levels" available for distribution in the USA and Canada. As mentioned previously, using the dictionary file in **Table 2** which contains all the possible drug names in this study, the drug names are automatically compared one by one and (since drugs may have different generic or trade names) then replaced by the most frequent name. As indicated in the Preprocessing section, the same procedure is also used for consolidating the test names using another file containing the test names for the extracted tables. All of these are summarized for the tabular data in **Figure 2A** and the corresponding extracted fields in **Figure 2B**.

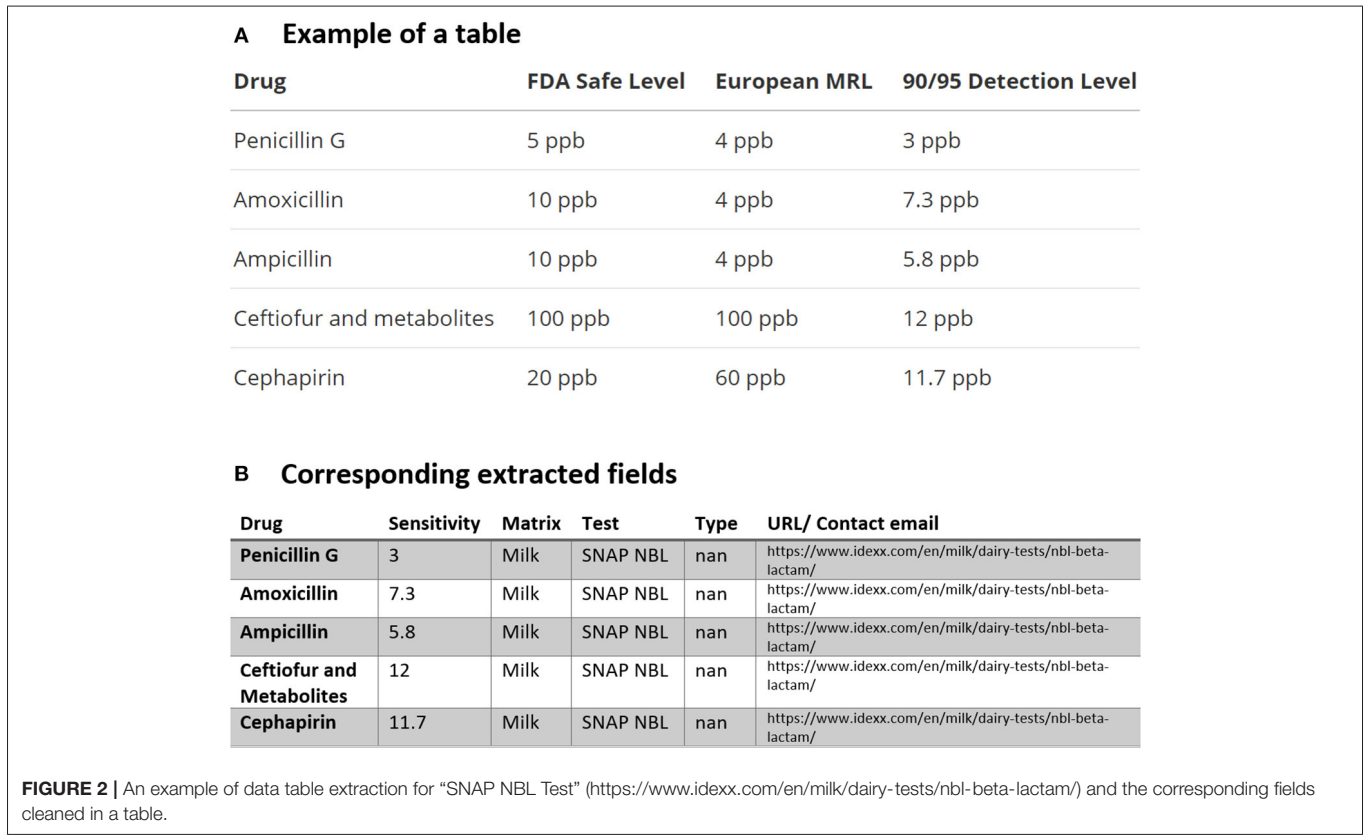

FIGURE 2 | An example of data table extraction for "SNAP NBL Test" (https://www.idexx.com/en/milk/dairy-tests/nbl-beta-lactam/) and the corresponding fields cleaned in a table.

## Extracting Information From Unstructured PDF Files

The main challenge is to deal with the extraction of data and information from PDF documents. Below we detail how we were able to collect data from different unstructured PDF files.

### Collecting the PDF Documents

The first step is to locate PDF documents containing information and fields of interest for parsing. In this study, seven manufacturers of commercially available rapid residue screening assay tests were contacted to obtain PDF files containing the specifications for available tests (1, 3). Once the PDFs were identified as potentially having useful information from these web sources to collect data, the documents are automatically stored for further analysis. It is worth mentioning that harvesting these dynamic web sources from one manufacturer (Charm Sciences, Inc.) is fast and straightforward in code implementation as these documents are put in a structured format on the Web based on the year, month, and test (as shown in **Figure 3**). Using *requests* and *BeautifulSoup* packages in Python, all the PDF files with the titles containing "MRK" by avoiding cases sensitivity of uppercase or lowercase for each letter (e.g., "mRK," "MrK," "mrk," etc.) from the years 2018, 2019, and 2020 were automatically collected and saved in a separate folder for further steps. In this process, we obtained 233 PDF files in total that may contain the necessary information for the rapid assay tests.

### Extracting Tables From PDFs

Deep learning tabular data was then implemented to retrieve data from each PDF obtained in the previous step. First, each document is consecutively investigated for tables (page by page) using a built-in module of Python called *read_pdf*, which is a function from a Python package called *tabula*. Once a table is found in a particular PDF, the desired keywords as presented in **Table 2** are searched in the document, and if any are found, the column names are then modified. Similar to the HTML documents, using the dictionary file in **Table 2**, each drug is automatically linked one after another one and (since drugs may have different generic or trade names) then replaced by the most frequent name. In these PDFs, the test, the matrix, and the type are not among the column names. Therefore, the text extracting *PyPDF2*, a library built as a PDF toolkit and is a function of the Python package *PdfFileReader,* is used to read the PDF containing tables and search for the three fields. These three fields, along with sensitivity and the page source of each PDF are saved in an excel file for later comparison with the original Excel file.

Below we provided three examples of tabular data that were that parsed and analyzed for data extraction. The first typical example of tables found in multiple PDFs is demonstrated in **Figure 4A**. Our PDF parser was able to clean and retrieve data as shown in the corresponding extracted table in **Figure 4B** Examples such as **Figure 4** require the least software-intensive method as all the information in rows of these types of tables is successfully extracted. However, we have detected other cases where one or more rows of the table are missing after data collection. For instance, **Figure 5B** shows the extracted tables of **Figure 5A**. As can be seen, the last row is missing. To fix this issue, we needed to revise our code, search for cases similar to **Figure 5** and then redo the data extraction to find missing rows and then fill the extracted documents in a semi-automatic way.

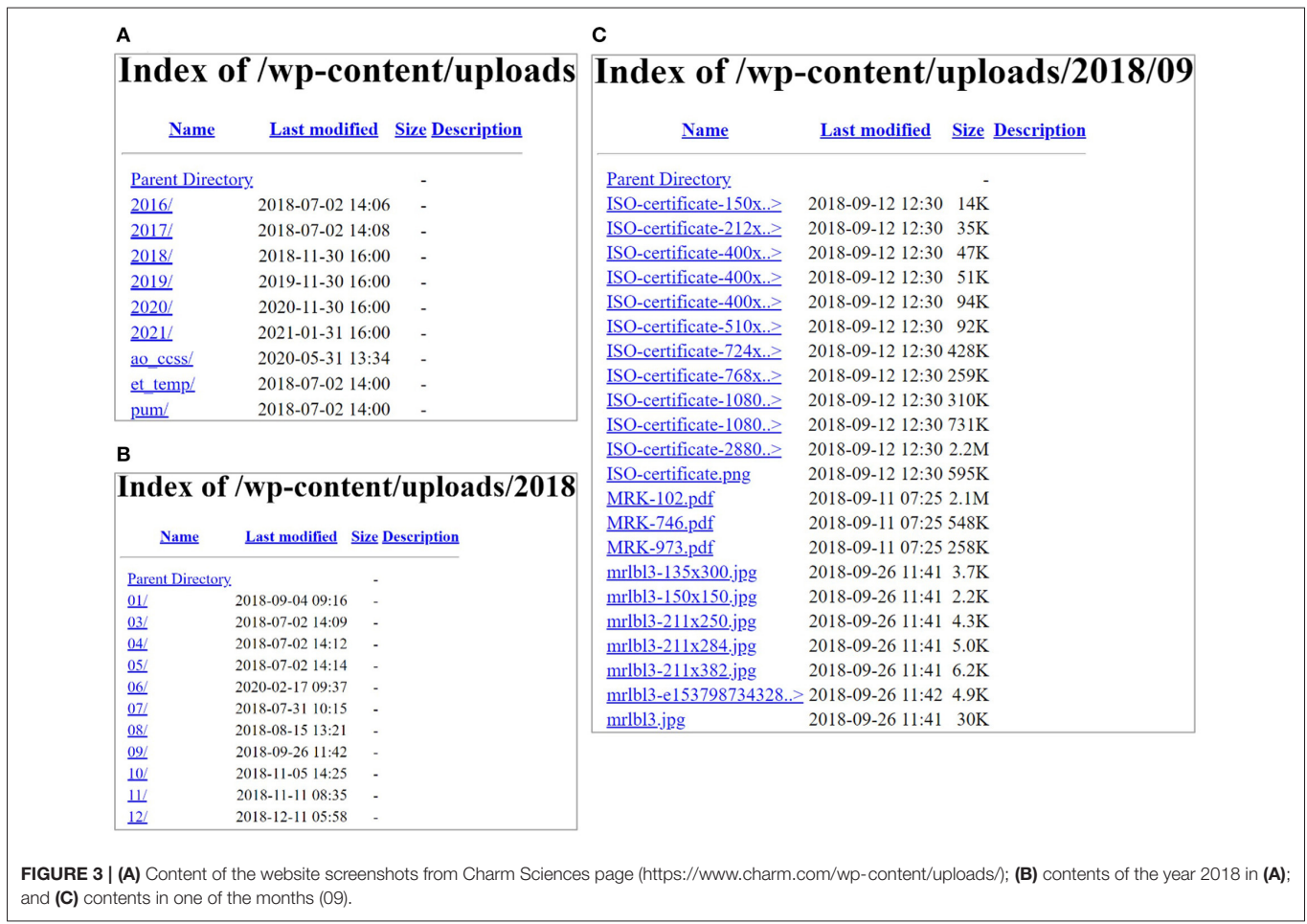

**FIGURE 3 | (A)** Content of the website screenshots from Charm Sciences page (https://www.charm.com/wp-content/uploads/); **(B)** contents of the year 2018 in **(A)**; and **(C)** contents in one of the months (09).

Another issue faced was related to tables containing multiple titles for different drug classes. These types of tables when parsed created confusion for our text mining model and treated a table with multiple titles similar to the example shows below in **Figure 6A**. As can be seen, another row exists in the middle of the table for a different drug class with the title. Our first assumption was each table may have only one title, and we searched for the unique title to extract information for drugs or rapid assay tests from that specific title. Since we did not emphasize much on any drug classes in this study, these rows are simply removed by checking if rows only contain strings and do not have any positive real numbers in the entire row. Other unnecessary information was also removed from tables including the headlines as shown by the blue ribbon in **Figure 6A**. We then obtained the corresponding extracted table as given by **Figure 6B**. It is worth pointing out that one row is also missing and needed further investigation.

Another case that we had to deal with was related to tables that had repeated column headings or tables that were broken into two sub-tables located side by side. As an example, we can observe in **Figure 7A** that for each field there are two headings for antimicrobial drugs, sensitivity (concentration in ppb), etc. Since in the PDF documents, such tables are always present, we had to code in a way to check the possibility of repeated columns for all the extracted tables. If any are found, the repeated columns are merged into a single column for each field. Therefore, the result of the corresponding extracted table is shown in **Figure 7B**.

Last but not least, it is worth mentioning that we also had cases where it was a combination of **Figures 6**, **7**. We then had to check these cases and ensure to capture all the available data from the PDF files. As explained above, since such cases are available in our files, we needed to check all of these scenarios one by one for each and every single document.

**A** Example of a table

**Sensitivity in Milk**

| Sensitivity Levels in Milk | | |
|---|---|---|
| Beta-lactam Drug | Charm 3 SL3 Sensitivity (ppb[A]) | Safe Level/Tolerance (ppb[A]) |
| Amoxicillin | 8.4 ppb | 10 ppb |
| Ampicillin | 8.0 ppb | 10 ppb |
| Ceftiofur[B] | 79 ppb | 100 ppb |
| Cephapirin | 20 ppb | 20 ppb |
| Cloxacillin | 8.6 ppb | 10 ppb |
| Penicillin G | 3.8 ppb | 5 ppb |

**B** Corresponding extracted fields

| Drug | Sensitivity | Matrix | Test | Type | URL/ Contact email |
|---|---|---|---|---|---|
| Amoxicillin | 8.4 ppb | Milk | Charm 3 SL3 Beta-Lactam | nan | https://www.charm.com/wp-content/uploads/2018/06/MRK-288.pdf |
| Ampicillin | 8.0 ppb | Milk | Charm 3 SL3 Beta-Lactam | nan | https://www.charm.com/wp-content/uploads/2018/06/MRK-288.pdf |
| Ceftiofur | 79 ppb | Milk | Charm 3 SL3 Beta-Lactam | nan | https://www.charm.com/wp-content/uploads/2018/06/MRK-288.pdf |
| Cephapirin | 20 ppb | Milk | Charm 3 SL3 Beta-Lactam | nan | https://www.charm.com/wp-content/uploads/2018/06/MRK-288.pdf |
| Cloxacillin | 8.6 ppb | Milk | Charm 3 SL3 Beta-Lactam | nan | https://www.charm.com/wp-content/uploads/2018/06/MRK-288.pdf |

**FIGURE 4** | An example of data table extraction for "Charm Flunixin and Beta-Lactam" and the corresponding fields cleaned in a table.

## Comparing the Extracted Tables With the Original Excel File

The last step is to consolidate the extracted information from tables in PDFs and the webpage sources with the original Excel file. For this purpose, any of the rows corresponding to the same drug with the same matrix, type, and a different sensitivity value will be updated. As a result, the sensitivity field is updated since new information was found in this case. If any of the three fields (drug, matrix, or type) are different, a new row with the new information will be added to the master Excel file. Following consolidation, this information was then uploaded to a publicly available online searchable interface (https://cafarad.ucdavis.edu/RapidAssay/rapidassay.aspx).

## DISCUSSION

Fields of natural language processing (NLP) and text mining provide tools and methodologies to retrieve relevant information (27, 28). However, most of the current strategies are limited to articles' textual body, usually ignoring tables and other formats of information, including figures. Tables can be hard to understand, even human readers struggle to understand the information (29, 30). Thus, the reader is required to consider a mental operation to obtain all the necessary information (30). Tables are used for other purposes, where authors need to present a relatively large amount of multi-dimensional information in a compact manner (31). Also, tables contain essential information needed for reproducibility of research and comparison to other studies. In addition to NLP, the text mining approach to tables performs poorly, and it is hard to understand the information that the table introduces. Information extraction from tables requires a multi-dimensional approach that will include pragmatic processing, syntactic processing and extraction, functional processing, and semantic tagging.

In our study, most of the files were produced and made available online by seven manufacturers in PDF formats while other tables were presented by HTML documents on Web sources. The latter were accurately abridged in an embodiment essential and accessible for our data collection system and provided a content management portal for interactive access to an encoded information reader system. The reconfigurable data collection process was arranged to be responsive with less challenge and configured data were expressed in an extensible markup language since the information on web pages is typically structured and thus, extracting tables and other desired information from it is straightforward.

The main challenge in extracting information was, however, to deal with PDF documents, which are not most often compressed or labeled for a reconfigurable data collection system and semantic parsing data collection, and do not follow any specific logical structural information (32). In this case, the relevant information is present in a table format which is typically unidentifiable after the PDF to text conversion. Due to the technical issues, any query on these documents relies upon accurate conversion of PDF documents to text, but long

**A** **Sensitivity in Milk**

| Sensitivity – Charm Flunixin and Beta-lactam Test ||||
| --- | --- | --- |
| Antibiotics and Veterinary Drugs | Sensitivity in Milk (ppb[A]) | FDA Safe Level/Tolerance (ppb[A]) |
| Amoxicillin | 5.9 | 10 |
| Ampicillin | 6.8 | 10 |
| Ceftiofur[B] | 63 | 100 |
| Cephapirin | 13.4 | 20 |
| Penicillin G | 2 | 5 |
| 5-hydroxyflunixin[C] Non-steroidal, Anti-imflammatory | 1.9 | 2 |

**B** **Corresponding extracted fields**

| Drug | Sensitivity | Matrix | Test | Type | URL/ Contact email |
| --- | --- | --- | --- | --- | --- |
| **Amoxicillin** | 5.9 | Milk | Charm Flunixin and Beta-Lactam | nan | https://www.charm.com/wp-content/uploads/2018/06/MRK-338.pdf |
| **Ampicillin** | 6.8 | Milk | Charm Flunixin and Beta-Lactam | nan | https://www.charm.com/wp-content/uploads/2018/06/MRK-338.pdf |
| **Ceftiofur** | 63 | Milk | Charm Flunixin and Beta-Lactam | nan | https://www.charm.com/wp-content/uploads/2018/06/MRK-338.pdf |
| **Cephapirin** | 13.4 | Milk | Charm Flunixin and Beta-Lactam | nan | https://www.charm.com/wp-content/uploads/2018/06/MRK-338.pdf |
| **Penicillin G** | 2 | Milk | Charm Flunixin and Beta-Lactam | nan | https://www.charm.com/wp-content/uploads/2018/06/MRK-338.pdf |
| **5-hydroxyflunixinC** Non-steroidal, Anti-inflammatory | 1.9 | Milk | Charm Flunixin and Beta-Lactam | nan | https://www.charm.com/wp-content/uploads/2018/06/MRK-338.pdf |

**FIGURE 5 |** An example of data table extraction for "Charm 3 SL3 Beta-Lactam" and the corresponding fields cleaned in a table where the information for Penicillin G in the last row was not extracted initially from our data collection.

lists of information in tables or inadvertent run-on sentences can lead to the erroneous determination of searched fields. If possible, prospects for correcting these technical malfunctions include revising the search engine model, improving algorithms to improve regular expression matching and learning for information extraction, and if available, obtaining documents in XML/HTML/JSON format to enhance PDF-to-text conversion.

It should also be noted that out of the total 233 PDFs, only 60 tables had relevant information and fields for data extraction. Out of 60 PDF documents, we were able to extract data points from 34 files providing approximately 1,100 records from commercial rapid assay tests. The other PDFs either did not have tables, or the extracted tables did not have relevant fields, or the tables could not be extracted in the first place. So, further investigation is needed, and it should be considered for the continuation of this research.

The data extracted in the study were made available to veterinarians and producers via an online interface (https://cafarad.ucdavis.edu/RapidAssay/rapidassay.aspx). Previously, these types of data were variably accessible for individual tests through company websites and in package inserts, but there was no centralized resource that veterinarians could refer to for information on tests available for a particular drug residue, and in some cases, the information could only be obtained through special requests to the manufacturer. In addition, manufacturer-provided information often does not include the FDA-approved tolerance values for that drug residue, thus making it difficult to know whether the commercial rapid assay test detects residues down to or below the tolerance. If there is a paucity of scientific data on which to base a withdrawal interval after extra-label drug use, it is often necessary to test for drug residues once the estimated withdrawal interval has been observed, prior to returning a treated animal or its products to food production. In the dairy industry particularly, producers and veterinarians often test the milk of all cattle that have been treated with a drug (whether label or extra-label drug use), after observing the required withdrawal interval and before returning the cow to the lactating herd. Therefore, a single online reference source where information on species, matrix, assay sensitivity, and FDA-approved tolerances for these rapid assay tests is a valuable resource for food animal veterinarians.

Overall, automatically collecting data from web pages and updating the corresponding data in the current available excel file resulted in the following advantages: (i) obtaining the sensitivity value corresponding to a specific test for a drug and a matrix without conducting the test, (ii) decreasing the errors caused by manually collecting and inserting the data, (iii) decreasing time and cost of obtaining sensitivity values since this will not be dependent on people for manually extracting the documents and data, (iv) more documents can be investigated for useful information, and (v) real-time implementation of the text mining for dynamic web sources is the most advantage of such model

**A** **Example of a table**

**Sensitivity in Milk**

| Detection Ranges in Cow Milk at 0 to 7 °C | | |
|---|---|---|
| Beta-lactam Drug | Detection Range (ppb$^A$) | EU/CODEX MRL (ppb$^A$) |
| Amoxicillin | 2.5 to 4 | 4 |
| Ampicillin | 2.5 to 4 | 4 |
| Cefacetrile | 6 to 12 | 125 |
| Cefalexin | 15 to 30 | 100 |
| Cefalonium | 3 to 5 | 20 |
| Cefazolin | 8 to 16 | 50 |
| Cefoperazone | 4 to 8 | 50 |
| Cefquinome | 15 to 20 | 20 |
| Ceftiofur and Metabolite$^B$ | 10 to 20 | 100/100 |
| Cephapirin | 4 to 8 | 60$^C$ |
| Cloxacillin | 25 to 35 | 30 |
| Dicloxacillin | 20 to 30 | 30 |
| Penicillin G | 2 to 3 | 4/4 |
| Tetracycline Drug | Detection Ranges (ppb$^A$) | EU/CODEX MRL (ppb$^A$) |
| Chlortetracycline | 50 to 100 | 100/100 |
| Oxytetracycline | 50 to 100 | 100/100 |
| Tetracycline | 10 to 30 | 100 |

**B** **Correspoding extracted fields**

| Drug | Sensitivity | Matrix | Test | Type | URL/ Contact email |
|---|---|---|---|---|---|
| **Amoxicillin** | 2.5 to 4 | Milk | Charm MRL Beta-Lactam | nan | https://www.charm.com/wp-content/uploads/2018/06/MRK-210_email.pdf |
| **Ampicillin** | 2.5 to 4 | Milk | Charm MRL Beta-Lactam | nan | https://www.charm.com/wp-content/uploads/2018/06/MRK-210_email.pdf |
| **Cefacetrile** | 6 to 12 | Milk | Charm MRL Beta-Lactam | nan | https://www.charm.com/wp-content/uploads/2018/06/MRK-210_email.pdf |
| **Cefalexin** | 15 to 30 | Milk | Charm MRL Beta-Lactam | nan | https://www.charm.com/wp-content/uploads/2018/06/MRK-210_email.pdf |
| **Cefalonium** | 3 to 5 | Milk | Charm MRL Beta-Lactam | nan | https://www.charm.com/wp-content/uploads/2018/06/MRK-210_email.pdf |
| **Cefazolin** | 8 to 16 | Milk | Charm MRL Beta-Lactam | nan | https://www.charm.com/wp-content/uploads/2018/06/MRK-210_email.pdf |
| **Cefoperazone** | 4 to 8 | Milk | Charm MRL Beta-Lactam | nan | https://www.charm.com/wp-content/uploads/2018/06/MRK-210_email.pdf |
| **Cefquinome** | 15 to 20 | Milk | Charm MRL Beta-Lactam | nan | https://www.charm.com/wp-content/uploads/2018/06/MRK-210_email.pdf |
| **Ceftiofur and Metabolite** | 10 to 20 | Milk | Charm MRL Beta-Lactam | nan | https://www.charm.com/wp-content/uploads/2018/06/MRK-210_email.pdf |
| **Cephapirin** | 4 to 8 | Milk | Charm MRL Beta-Lactam | nan | https://www.charm.com/wp-content/uploads/2018/06/MRK-210_email.pdf |
| **Cloxacillin** | 25 to 35 | Milk | Charm MRL Beta-Lactam | nan | https://www.charm.com/wp-content/uploads/2018/06/MRK-210_email.pdf |
| **Dicloxacillin** | 20 to 30 | Milk | Charm MRL Beta-Lactam | nan | https://www.charm.com/wp-content/uploads/2018/06/MRK-210_email.pdf |
| **Penicillin G** | 2 to 3 | Milk | Charm MRL Beta-Lactam | nan | https://www.charm.com/wp-content/uploads/2018/06/MRK-210_email.pdf |
| Chlortetracycline | 50 to 100 | Milk | Charm MRL Beta-Lactam | nan | https://www.charm.com/wp-content/uploads/2018/06/MRK-210_email.pdf |
| Oxytetracycline | 50 to 100 | Milk | Charm MRL Beta-Lactam | nan | https://www.charm.com/wp-content/uploads/2018/06/MRK-210_email.pdf |

**FIGURE 6** | An example of data table extraction for "Charm MRL Beta-Lactam" and the corresponding fields cleaned in a table where the last row for Tetracycline was not extracted initially from our data collection.

development. This has a real-world veterinary application as being able to automatically collect and update residue assay tests allows for access to up-to-date information that helps drive decision making regarding which tests to use, whose results that can then help determine if food products will be safe for human consumption.

### A  Example of a table

**Sensitivity in Milk**

**Sensitivity and Selectivity**

**Selectivity** - Antimicrobial drug-free samples should yeild 90% negative results with 95% confidence.
**Sensitivity** - Antimicrobial drugs detected as positive compared to regulatory levels.

| Antimicrobial Drug[A] | Concentration[B] (ppb[C]) | US Safe Level/ Tolerance (ppb[C]) | EU/CODEX MRL[D] (µg/kg) | Antimicrobial Drug[A] | Concentration[B] (ppb[C]) | US Safe Level/ Tolerance (ppb[C]) | EU/CODEX MRL[D] (µg/kg) |
|---|---|---|---|---|---|---|---|
| Amoxicillin | 3 to 4 | 10 | 4/4 | Gentamicin | 75 to 150 | 30 | 100 / 200 |
| Ampicillin | 3 to 4 | 10 | 4 | Lincomycin | 75 to 150 | None | 150 |
| Cefacetrile | 10 to 15 | None | 125 | Nafcillin | 5 to 10 | None | 30 |
| Cefalexin | 75 to 100 | None | 100 | Neomycin | 100 to 150 | 150 | 1500 / 1500 |
| Cefalonium | 15 to 20 | None | 20 | Oxacillin | 5 to 10 | None | 30 |
| Cefazolin | 6 to 10 | None | 50 | Oxytetracycline | 75 to 100 | 300 | 100 / 100 |
| Cefoperazone | 20 to 30 | None | 50 | Penethamate[G] | 2 to 3 | None | 4 |
| Cefquinome | 40 to 60 | None | 20 | Penicillin G | 2 to 3 | 5 | 4 / 4 |
| Ceftiofur & Metabolites[E] | 50 to 100 | 100 | 100 / 100 | Pirlimycin | 25 to 50 | None | 100/100 |
| Cefuroxime | 20 to 25 | None | None | Spiramycin | 300 to 400 | None | 200 / 200 |
| Cephapirin | 8 to 10 | 20 | 60 | Sulfadiazine | 40 to 60 | 10 | 100 |
| Chlortetracycline | 200 to 300 | 300 | 100/100 | Sulfadimethoxine | 25 to 50 | 10 | 100 |
| Cloxacillin | 10 to 25 | 10 | 30 | Sulfamethazine (Sulfadimidine) | 75 to 125 | 10 | 100 / 25 |
| Dapsone | 1 to 2 | None | 0[F] | Tetracycline | 50 to 100 | 300 | 100 / 100 |
| Dicloxacillin | 5 to 10 | None | 30 | Tilmicosin | 25 to 35 | None | 50 |
| Doxycycline | 25 to 75 | None | 0[F] | Trimethoprim | 200 to 300 | None | 50 |
| Erythromycin | 75 to 100 | 50 | 40 | Tylosin | 20 to 30 | 50 | 50 |

### B  Corresponding extracted fields

| Drug | Sensitivity | Matrix | Test | Type | URL/ Contact email |
|---|---|---|---|---|---|
| **Amoxicillin** | 3 to 4 | Milk | Charm Cowside II | | https://www.charm.com/wp-content/uploads/2018/06/MRK-003.pdf |
| **Ampicillin** | 3 to 4 | Milk | Charm Cowside II | | https://www.charm.com/wp-content/uploads/2018/06/MRK-003.pdf |
| **Cefacetrile** | 10 to 15 | Milk | Charm Cowside II | | https://www.charm.com/wp-content/uploads/2018/06/MRK-003.pdf |
| **Cefalexin** | 75 to 100 | Milk | Charm Cowside II | | https://www.charm.com/wp-content/uploads/2018/06/MRK-003.pdf |
| **Cefalonium** | 15 to 20 | Milk | Charm Cowside II | | https://www.charm.com/wp-content/uploads/2018/06/MRK-003.pdf |
| **Cefazolin** | 6 to 10 | Milk | Charm Cowside II | | https://www.charm.com/wp-content/uploads/2018/06/MRK-003.pdf |
| **Cefoperazone** | 20 to 30 | Milk | Charm Cowside II | | https://www.charm.com/wp-content/uploads/2018/06/MRK-003.pdf |
| **Cefquinome** | 40 to 60 | Milk | Charm Cowside II | | https://www.charm.com/wp-content/uploads/2018/06/MRK-003.pdf |
| **Ceftiofur & Metabolites** | 50 to 100 | Milk | Charm Cowside II | | https://www.charm.com/wp-content/uploads/2018/06/MRK-003.pdf |
| **Cefuroxime** | 20 to 25 | Milk | Charm Cowside II | | https://www.charm.com/wp-content/uploads/2018/06/MRK-003.pdf |
| **Cephapirin** | 8 to 10 | Milk | Charm Cowside II | | https://www.charm.com/wp-content/uploads/2018/06/MRK-003.pdf |
| **Chlortetracycline** | 200 to 300 | Milk | Charm Cowside II | | https://www.charm.com/wp-content/uploads/2018/06/MRK-003.pdf |
| **Cloxacillin** | 10 to 25 | Milk | Charm Cowside II | | https://www.charm.com/wp-content/uploads/2018/06/MRK-003.pdf |
| **Dapsone** | 1 to 2 | Milk | Charm Cowside II | | https://www.charm.com/wp-content/uploads/2018/06/MRK-003.pdf |
| **Dicloxacillin** | 5 to 10 | Milk | Charm Cowside II | | https://www.charm.com/wp-content/uploads/2018/06/MRK-003.pdf |
| **Doxycycline** | 25 to 75 | Milk | Charm Cowside II | | https://www.charm.com/wp-content/uploads/2018/06/MRK-003.pdf |
| **Gentamicin** | 75 to 150 | Milk | Charm Cowside II | | https://www.charm.com/wp-content/uploads/2018/06/MRK-003.pdf |
| **Lincomycin** | 75 to 150 | Milk | Charm Cowside II | | https://www.charm.com/wp-content/uploads/2018/06/MRK-003.pdf |
| **Nafcillin** | 5 to 10 | Milk | Charm Cowside II | | https://www.charm.com/wp-content/uploads/2018/06/MRK-003.pdf |
| **Neomycin** | 100 to 150 | Milk | Charm Cowside II | | https://www.charm.com/wp-content/uploads/2018/06/MRK-003.pdf |
| **Oxacillin** | 5 to 10 | Milk | Charm Cowside II | | https://www.charm.com/wp-content/uploads/2018/06/MRK-003.pdf |
| **Oxytetracycline** | 75 to 100 | Milk | Charm Cowside II | | https://www.charm.com/wp-content/uploads/2018/06/MRK-003.pdf |
| **Penethamate** | 2 to 3 | Milk | Charm Cowside II | | https://www.charm.com/wp-content/uploads/2018/06/MRK-003.pdf |
| **Penicillin G** | 2 to 3 | Milk | Charm Cowside II | | https://www.charm.com/wp-content/uploads/2018/06/MRK-003.pdf |
| **Pirlimycin** | 25 to 50 | Milk | Charm Cowside II | | https://www.charm.com/wp-content/uploads/2018/06/MRK-003.pdf |
| **Spiramycin** | 300 to 400 | Milk | Charm Cowside II | | https://www.charm.com/wp-content/uploads/2018/06/MRK-003.pdf |
| **Sulfadiazine** | 40 to 60 | Milk | Charm Cowside II | | https://www.charm.com/wp-content/uploads/2018/06/MRK-003.pdf |
| **Sulfadimethoxine** | 25 to 50 | Milk | Charm Cowside II | | https://www.charm.com/wp-content/uploads/2018/06/MRK-003.pdf |
| **Sulfadimidine** | 75 to 125 | Milk | Charm Cowside II | | https://www.charm.com/wp-content/uploads/2018/06/MRK-003.pdf |
| **Tetracycline** | 50 to 100 | Milk | Charm Cowside II | | https://www.charm.com/wp-content/uploads/2018/06/MRK-003.pdf |
| **Tilmicosin** | 25 to 35 | Milk | Charm Cowside II | | https://www.charm.com/wp-content/uploads/2018/06/MRK-003.pdf |
| **Trimethoprim** | 200 to 300 | Milk | Charm Cowside II | | https://www.charm.com/wp-content/uploads/2018/06/MRK-003.pdf |

**FIGURE 7** | An example of data table extraction for "Charm Cowside II" and the corresponding fields cleaned in a table.

Unstructured data is the accessible form of data and mostly presented in the form of publications that can be generated in any application scenario. Therefore, there has been an extreme necessity to devise methods and algorithms that are capable of efficiently processing an extensive variety of text applications from electronic documents. This study has provided a data-mining method for automatically extracting rapid assay data of the different document types which are common in the text domain, with a distinct target of table extraction. However, data are not always presented in the form of tables. As part of our future work, it is worth mentioning that we are currently developing and working on a learned information extraction system to transform any text format of pharmacokinetic data into more structured data which is then mined for the use of therapeutic drugs in modern animal agriculture including recommendations for safe withdrawal intervals of drugs and chemicals in food-producing animals. The objective of the data mining project is to create a comprehensive drug database to help the mission of FARAD by improving overall animal health and promoting efficient and humane production practices.

## DATA AVAILABILITY STATEMENT

The original contributions presented in the study are included in the article/**Supplementary Material**, further inquiries can be directed to the corresponding author/s.

## AUTHOR CONTRIBUTIONS

MJ-D: conceptualization, methodology, validation, writing—original draft, writing—review, editing, funding acquisition, and data science. ST: model development, validation, resources, writing—draft, writing—review, and editing. NM: conceptualization, writing—review, editing, online repository, and data science. XX: conceptualization, software, writing—review, editing, and data science. ER and FM: writing—review, editing, and interpretation. NZ: writing—review and editing. LT: conceptualization, methodology, writing—review, editing, and funding acquisition. All authors contributed to the article and approved the submitted version.

## FUNDING

This work was supported by the USDA via the FARAD program and its support for the 1DATA Consortium at Kansas State University. MJ-D also accepted funding from BioNexus KC for this project. Neither USDA nor BioNexus KC had direct role in this article.

## SUPPLEMENTARY MATERIAL

The Supplementary Material for this article can be found online at: https://www.frontiersin.org/articles/10.3389/fvets.2021.674730/full#supplementary-material